\documentclass[10pt,twocolumn]{article}

\usepackage[dvipsnames]{xcolor}
\usepackage{cvpr}
\usepackage{times}
\usepackage{epsfig}
\usepackage{graphicx}
\usepackage{amsmath}
\usepackage{amssymb}
\usepackage{bm}
\usepackage{algorithm}
\usepackage[noend]{algorithmic}
\usepackage{subcaption}
\usepackage{gensymb}
\usepackage{pifont}
\usepackage{tabulary}

\newcolumntype{K}[1]{>{\centering\arraybackslash}p{#1}}

\newcommand{\cmark}{\Large \color{ForestGreen}\ding{51}}%
\newcommand{\xmark}{\Large \color{red}\ding{55}}%

\newcommand{\figref}[1]{Fig.~\ref{fig:#1}}
\newcommand{\tblref}[1]{Table~\ref{tbl:#1}}
\newcommand{\algref}[1]{Alg.~\ref{alg:#1}}
\newcommand{\secref}[1]{Sec.~\ref{sec:#1}}

\def\real{\mathbb{R}}
\newcommand{\squeeze}{\vspace{0mm}}

\usepackage[pagebackref=true,breaklinks=true,letterpaper=true,colorlinks,bookmarks=false]{hyperref}

\cvprfinalcopy 

\def\cvprPaperID{2461} 

\ifcvprfinal\fi
\begin{document}

\title{Compact Bilinear Pooling}

\author{
Yang Gao$^{1}$, \;  Oscar Beijbom$^1$, \; Ning Zhang$^2$\thanks{This work was done when Ning Zhang was in Berkeley.}, \; Trevor Darrell$^1$ \thanks{Prof. Darrell was supported in part by DARPA; AFRL; DoD MURI award N000141110688; NSF awards  IIS-1212798, IIS-1427425, and IIS-1536003, and the Berkeley Vision and Learning Center.} \\
$^1$EECS, UC Berkeley ~~~~~~~~~~
$^2$Snapchat Inc.\\
{\tt\small \{yg, obeijbom, trevor\}@eecs.berkeley.edu} \:\:\:\:
{\tt\small \{ning.zhang\}@snapchat.com}
}

\maketitle

\begin{abstract}
Bilinear models has been shown to achieve impressive performance on a wide range of visual tasks, such as semantic segmentation, fine grained recognition and face recognition. However, bilinear features are high dimensional, typically on the order of hundreds of thousands to a few million, which makes them impractical for subsequent analysis. We propose two compact bilinear representations with the same discriminative power as the full bilinear representation but with only a few thousand dimensions. Our compact representations allow back-propagation of classification errors enabling an end-to-end optimization of the visual recognition system. The compact bilinear representations are derived through a novel kernelized analysis of bilinear pooling which provide insights into the discriminative power of bilinear pooling, and a platform for further research in compact pooling methods. Experimentation illustrate the utility of the proposed representations for image classification and few-shot learning across several datasets.
\end{abstract}

\section{Introduction}
Encoding and pooling of visual features is an integral part of semantic image analysis methods. Before the influential 2012 paper of Krizhevsky et al.~\cite{krizhevsky2012imagenet} rediscovering the models pioneered by \cite{lecun} and related efforts, such methods typically involved a series of independent steps: feature extraction, encoding, pooling and classification; each thoroughly investigated in numerous publications as the bag of visual words (BoVW) framework. Notable contributions include HOG~\cite{dalal2005histograms}, and SIFT~\cite{lowe1999object} descriptors, fisher encoding~\cite{perronnin2010improving}, bilinear pooling~\cite{carreira2012semantic} and spatial pyramids~\cite{lazebnik2006beyond}, each significantly improving the recognition accuracy. 

\begin{figure}[t]
\begin{center}
   \includegraphics[width=\linewidth]{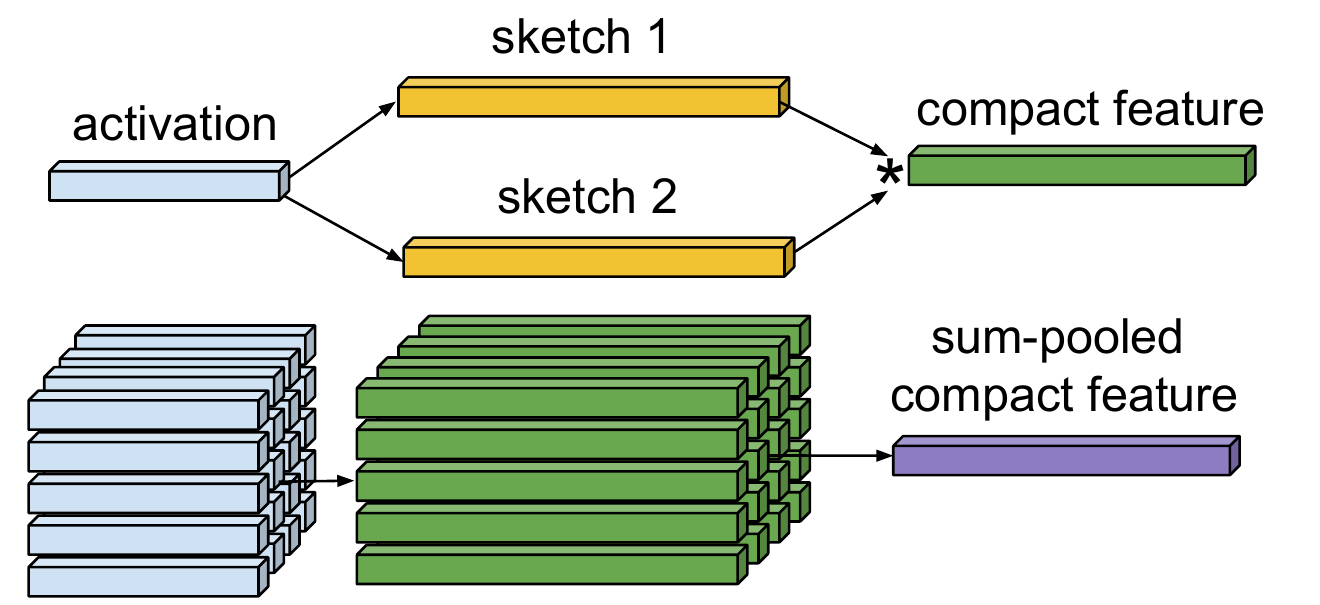}
\end{center}
   \caption{We propose a compact bilinear pooling method for image classification. Our pooling method is learned through end-to-end back-propagation and enables a low-dimensional but highly discriminative image representation. Top pipeline shows the Tensor Sketch projection applied to the activation at a single spatial location, with $*$ denoting circular convolution. Bottom pipeline shows how to obtain a global compact descriptor by sum pooling.}
\label{fig:system}
\squeeze
\end{figure}

Recent results have showed that end-to-end back-propagation of gradients in a convolutional neural network (CNN) enables joint optimization of the whole pipeline, resulting in significantly higher recognition accuracy. While the distinction of the steps is less clear in a CNN than in a BoVW pipeline, one can view the first several convolutional layers as a feature extractor and the later fully connected layers as a pooling and encoding mechanism. This has been explored recently in methods combining the feature extraction architecture of the CNN paradigm, with the pooling \& encoding steps from the BoVW paradigm~\cite{lin2015bilinear, cimpoi2015deep}. Notably, Lin et al. recently replaced the fully connected layers with bilinear pooling achieving remarkable improvements for fine-grained visual recognition~\cite{lin2015bilinear}. However, their final representation is very high-dimensional; in their paper the encoded feature dimension, $d$, is more than $250,000$. Such representation is impractical for several reasons: (1) if used with a standard one-vs-rest linear classifier for $k$ classes, the number of model parameters becomes $kd$, which for e.g. $k=1000$ means $>250$ million model parameters, (2) for retrieval or deployment scenarios which require features to be stored in a database, the storage becomes expensive; storing a millions samples requires 2TB of storage at double precision, (3) further processing such as spatial pyramid matching~\cite{lazebnik2006beyond}, or domain adaptation~\cite{daume2009frustratingly} often requires feature concatenation; again, straining memory and storage capacities, and (4) classifier regularization, in particular  under few-shot learning scenarios becomes challenging~\cite{fei2006one}. The main contribution of this work is a pair of bilinear pooling methods, each able to reduce the feature dimensionality three orders of magnitude with little-to-no loss in performance compared to a full bilinear pooling. The proposed methods are motivated by a novel kernelized viewpoint of bilinear pooling, and, critically, allow back-propagation for end-to-end learning. 

Our proposed compact bilinear methods rely on the existence of low dimensional feature maps for kernel functions. Rahimi \cite{rahimi2007random} first proposed a method to find explicit feature maps for Gaussian and Laplacian kernels. This was later extended for the intersection kernel, $\chi^2$ kernel and the exponential $\chi^2$ kernel~\cite{vedaldi2012efficient,maji2009max,vempati2010generalized}. We show that bilinear features are closely related to polynomial kernels and propose new methods for compact bilinear features based on algorithms for the polynomial kernel first proposed by Kar \cite{kar2012random} and Pham \cite{pham2013fast}; a key aspect of our contribution is that we show how to back-propagate through such representations. 

\textbf{Contributions:} The contribution of this work is three-fold. First, we propose two compact bilinear pooling methods, which can reduce the feature dimensionality two orders of magnitude with little-to-no loss in performance compared to a full bilinear pooling. Second, we show that the back-propagation through the compact bilinear pooling can be efficiently computed, allowing end-to-end optimization of the recognition network. Third, we provide a novel kernelized viewpoint of bilinear pooling which not only motivates the proposed compact methods, but also provides theoretical insights into bilinear pooling. Implementations of the proposed methods, in Caffe and MatConvNet, are publicly available: \url{https://github.com/gy20073/compact_bilinear_pooling}

\squeeze
\section{Related work}
Bilinear models were first introduced by Tenenbaum and Freeman \cite{tenenbaum2000separating} to separate style and content. Second order pooling have since been considered for both semantic segmentation and fine grained recognition, using both hand-tuned~\cite{carreira2012semantic}, and learned features~\cite{lin2015bilinear}. Although repeatedly shown to produce state-of-the art results, it has not been widely adopted; we believe this is partly due to the prohibitively large dimensionality of the extracted features.

Several other clustering methods have been considered for visual recognition. Leung and Malik used vector quantization in the Bag of Visual Words (BoVW) framework~\cite{leung2001representing} initially used for texture classification, but later adopted for other visual tasks. VLAD \cite{jegou2010aggregating} and Improved Fisher Vector \cite{perronnin2010improving} encoding improved over hard vector quantization by including second order information in the descriptors. Fisher vector has been recently been used to achieved start-of-art performances on many data-sets\cite{cimpoi2015deep}. 

Reducing the number of parameters in CNN is important for training large networks and for deployment (e.g. on embedded systems). Deep Fried Convnets \cite{yang2014deep} aims to reduce the number of parameters in the fully connected layer, which usually accounts for 90\% of parameters. Several other papers pursue similar goals, such as the Fast Circulant Projection which uses a circular structure to reduce memory and speed up computation \cite{cheng2015fast}. Furthermore, Network in Network \cite{lin2013network} uses a micro network as the convolution filter and achieves good performance when using only global average pooling. We take an alternative approach and focus on improving the efficiency of bilinear features, which outperform fully connected layers in many studies \cite{carreira2012semantic,cimpoi2015deep,roychowdhury2015face}. 


\begin{table}
\small
\centering \begin{tabular}{p{1.3cm} | K{.6cm} K{1.5cm} K{1.3cm} K{1.5cm}}
& Com-pact & Highly discriminative & Flexible input size& End-to-end learnable\\
\hline
\hline
Fully connected & \cmark &  \xmark &\xmark  & \cmark \\
\hline
Fisher encoding & \xmark & \cmark & \cmark & \xmark \\
\hline
Bilinear pooling & \xmark & \cmark& \cmark& \cmark \\
\hline
Compact bilinear &\cmark & \cmark & \cmark& \cmark \\
\hline    
\end{tabular}
  \caption{Pooling methods property overview: Fully connected pooling \cite{krizhevsky2012imagenet}, is compact and can be learned end-to-end by back propagation, but it requires a fixed input image size and is less discriminative than other methods~\cite{cimpoi2015deep, lin2015bilinear}. Fisher encoding is more discriminative but high dimensional and can not be learned end-to-end~\cite{cimpoi2015deep}. Bilinear pooling is discriminative and tune-able but very high dimensional~\cite{lin2015bilinear}. Our proposed compact bilinear pooling is as effective as bilinear pooling, but much more compact.}
   \vskip -0.5cm
\label{tbl:properties}

\end{table}

\section{Compact bilinear models}
\label{sec:methods}
Bilinear pooling~\cite{lin2015bilinear} or second order pooling \cite{carreira2012semantic} forms a global image descriptor by calculating:
\begin{align}
B({\cal X}) = \sum_{s \in \cal S} x_s x^T_s  \label{eq:bldef}
\end{align}
where ${\cal X}=({x}_1, \hdots, {x}_{|{\cal S}|}, {x}_s \in \real^c)$ is a set of local descriptors, and ${\cal S}$ is the set of spatial locations (combinations of rows \& columns). Local descriptors, $x_s$ are typically extracted using SIFT\cite{lowe1999object}, HOG \cite{dalal2005histograms} or by a forward pass through a CNN~\cite{krizhevsky2012imagenet}. As defined in \eqref{eq:bldef}, $B({\cal X})$ is a $c\times c$ matrix, but for the purpose of our analysis, we will view it as a length $c^2$ vector.

\subsection{A kernelized view of bilinear pooling}
Image classification using bilinear descriptors is typically achieved using linear Support Vector Machines (SVM) or logistic regression. These can both be viewed as linear kernel machines, and we provide an analysis below\footnote{We ignore the normalization (signed square root and $\ell_2$ normalization) which is typically applied before classification}. Given two sets of local descriptors: ${\cal X}$ and ${\cal Y}$, a linear kernel machine compares these as:
\begin{align}
\begin{split}
\label{eq:bilinear_kernel}
\langle B({\cal X}), B({\cal Y}) \rangle &= \langle \sum_{s\in {\cal S}} x_s x_s^T, \sum_{u \in {\cal U}} y_u y_u^T \rangle \\
& = \sum_{s\in {\cal S}} \sum_{u \in {\cal U}} \langle x_s x_s^T,y_u y_u^T \rangle \\
&= \sum_{s\in {\cal S}} \sum_{u \in {\cal U}} \langle x_s, y_u \rangle ^2
\end{split}
\end{align}
From the last line in \eqref{eq:bilinear_kernel}, it is clear that the bilinear descriptor compares each local descriptor in the first image with that in the second image and that the comparison operator is a second order polynomial kernel. Bilinear pooling thus gives a linear classifier the discriminative power of a second order kernel-machine, which may help explain the strong empirical performance observed in previous work~\cite{lin2015bilinear,carreira2012semantic,cimpoi2015deep,roychowdhury2015face}.

\subsection{Compact bilinear pooling}
In this section we define the proposed compact bilinear pooling methods. Let $k(x, y)$ denote the comparison kernel, i.e. the second order polynomial kernel. If we could find some low dimensional projection function $\phi(x) \in R^d$, where $d<<c^2$, that satisfy $\langle \phi(x), \phi(y) \rangle \approx k(x, y)$, then we could approximate the inner product of \eqref{eq:bilinear_kernel} by: 
\begin{align}
\begin{split}
\label{eq:compact_bilinear_kernel}
\langle B({\cal X}), B({\cal Y}) \rangle &= \sum_{s\in {\cal S}} \sum_{u \in {\cal U}} \langle x_s, y_u \rangle^2\\
								& \approx \sum_{s\in {\cal S}} \sum_{u \in {\cal U}} \langle \phi(x), \phi(y) \rangle \\
                                &\equiv \langle C({\cal X}), C({\cal Y}) \rangle,
\end{split}
\end{align}
where 
\begin{align}
C({\cal X}):=\sum_{s\in {\cal S}} \phi(x_s)
\end{align}
is the compact bilinear feature. It is clear from this analysis that any low-dimensional approximation of the polynomial kernel can be used to towards our goal of creating a compact bilinear pooling method. We investigate two such approximations: Random Maclaurin (RM)~\cite{kar2012random} and Tensor Sketch (TS)~\cite{pham2013fast}, detailed in \algref{RM} and \algref{TS} respectively. 

RM is an early approach developed to serve as a low dimensional explicit feature map to approximate the polynomial kernel~\cite{kar2012random}. The intuition is straight forward. If $w_1, w_2 \in \real^c$ are two random $-1, +1$ vectors and $\phi(x)=\langle w_1, x \rangle  \langle w_2, x \rangle$, then for non-random $x, y \in \real^c$, $E[\phi(x)\phi(y)] = E[\langle w_1, x\rangle \langle w_1, y\rangle ]^2=\langle x, y \rangle ^2$. Thus each projected entry in RM has an expectation of the quantity to be approximated. By using $d$ entries in the output, the estimator variance could be brought down by a factor of $1/d$. TS uses sketching functions to improve the computational complexity during projection and tend to provide better approximations in practice~\cite{pham2013fast}. Similar to the RM approach, Count Sketch\cite{charikar2002finding}, defined by $\Psi(x,h,s)$ in Algorithm \ref{alg:TS}, has the favorable property that: $E[\langle \Psi(x, h, s), \Psi(y, h, s) \rangle] = \langle x, y \rangle$ \cite{charikar2002finding}. Moreover, one can show that $\Psi(x \otimes y, h, s) = \Psi(x, h, s) * \Psi(y, h, s)$, i.e. the count sketch of two vectors' outer product is the convolution of individual's count sketch \cite{pham2013fast}. Then the same approximation in expectation follows.

\subsubsection{Back propagation of compact bilinear pooling}
\label{sec:compact_fine_tuning}
In this section we derive back-propagation for the two compact bilinear pooling methods and show they're efficient both in computation and storage.  

For RM, let $L$ denote the loss function, $s$ the spatial index, $d$ the projected dimension, $n$ the index of the training sample and $y_d^n \in \real$ the output of the RM layer at dimension $d$ for instance $n$. Back propagation of RM pooling can then be written as: \begin{align}
\begin{split}
\frac{\partial L}{\partial x_s^n}  &= \sum_d \frac{\partial L}{\partial y_d^n} \sum_k \langle W_k(d), x_s^n\rangle W_{\bar{k}}(d) \\
\frac{\partial L}{\partial W_k(d)} &= \sum_n \frac{\partial L}{\partial y_d^n} \sum_s \langle W_{\bar{k}}(d), x_s^n \rangle x_s^n
\end{split}
\end{align}
where $k=1,2$, $\bar{k}=2, 1$, and $W_k(d)$ is row $d$ of matrix $W_k$. For TS, using the same notation, 
\begin{align}
\begin{split}
\frac{\partial L}{\partial x_{s}^n} &= \sum_d \frac{\partial L}{\partial y_d^n} \sum_k T^k_{d}(x^n_s) \circ s_k \\
\frac{\partial L}{\partial s_k} &= \sum_{n,d} \frac{\partial L}{\partial y_{d}^n} \sum_s T^k_{d}(x^n_s) \circ x_{s}^n
\label{eq_TS}
\end{split}
\end{align}
where $T^k_d(x) \in R^c $ and $T^{k}_{d}(x)_c=\Psi(x, h_{\bar{k}}, s_{\bar{k}})_{d-h_k(c)}$. When $d-h_k(c)$ is negative, it denotes the circular index $(d-h_k(c))+D$, where $D$ is the projected dimensionality. Note that in TS, we could only get a gradient for $s_k$. $h_k$ is combinatorial, and thus fixed during back-prop. 

The back-prop equation for RM can be conveniently written as a few matrix multiplications. It has the same computational and storage complexity as its forward pass, and can be calculated efficiently. Similarly, Equation \ref{eq_TS} can also be expressed as a few FFT, IFFT and matrix multiplication operations. The computational and storage complexity of TS are also similar to its forward pass. 

\begin{algorithm}
   	\caption{Random Maclaurin Projection}
   	\label{alg:RM}
\begin{algorithmic}
   	\STATE Input: $x \in \real^c$
   	\STATE Output: feature map $\phi_{RM}(x) \in \real^d$,  such that $\langle \phi_{RM}(x), \phi_{RM}(y) \rangle \approx 
										\langle x, y \rangle ^2$
	
	\STATE 1. Generate random but fixed $W_1, W_2 \in \real^{d\times c}$, where each entry is either $+1$ or $-1$ with equal probability. 
	\STATE 2. Let $\phi_{RM}(x) \equiv \frac{1}{\sqrt{d}}(W_1 x) \circ (W_2 x)$, where $\circ$ denotes element-wise multiplication. 
\end{algorithmic}
\end{algorithm}

\begin{algorithm}
   	\caption{Tensor Sketch Projection}
   	\label{alg:TS}
\begin{algorithmic}
   	\STATE Input: $x \in \real^c$
   	\STATE Output: feature map $\phi_{TS}(x) \in \real^d$,  such that $\langle \phi_{TS}(x), \phi_{TS}(y) \rangle \approx 
										\langle x, y \rangle ^2$
	\STATE 1. Generate random but fixed $h_k \in \mathbb{N}^c$ and $s_k \in \{+1, -1\}^{c}$ where $h_k(i)$ is uniformly drawn from $\{1,2,\hdots, d\}$, $s_k(i)$ is uniformly drawn from $\{+1, -1\}$, and $k= 1, 2$. 
	\STATE 2. Next, define sketch function $\Psi(x, h, s)=\{(Qx)_1, \hdots , (Qx)_d\}$, where $(Qx)_j=\sum_{t: h(t)=j} s(t)x_t $
	\STATE 3. Finally, define $\phi_{TS}(x) \equiv \text{FFT}^{-1}(\text{FFT}(\Psi(x, h_1, s_1)) \circ \text{FFT}(\Psi(x, h_2, s_2)))$, where the $\circ$ denotes element-wise multiplication. 
\end{algorithmic}
\end{algorithm}



\begin{table*}
\small
\centering \begin{tabular}{l|ccc}
\hline
            & Full Bilinear        & Random Maclaurin (RM) & Tensor Sketch (TS)  \\
\hline\hline
Dimension       & $c^2$~[262K]    & $d$~[10K]             & $d$~[10K]            \\
Parameters Memory    & 0           & $2cd$~[40MB]          & $2c$~[4KB]          \\
Computation    & O($hwc^2$)    & O($hwcd$)           & O($hw(c+d\log d)$) \\
Classifier Parameter Memory & $kc^2$~[1000MB] & $kd$~[40MB]       & $kd$~[40MB]    \\
\hline    
\end{tabular}
\caption{Dimension, memory and computation comparison among bilinear and the proposed compact bilinear features. Parameters $c, d, h, w, k$ represent the number of channels before the pooling layer, the projected dimension of compact bilinear layer, the height and width of the previous layer and the number of classes respectively. Numbers in brackets indicate typical value when bilinear pooling is applied after the last convolutional layer of VGG-VD \cite{simonyan2014very} model on a 1000-class classification task, i.e. $c=512, d=10,000, h=w=13, k=1000$. All data are stored in single precision.} \label{tbl:theory_compare} \end{table*}

\subsubsection{Some properties of compact bilinear pooling}

\tblref{theory_compare} shows the comparison among bilinear and compact bilinear feature using RM and TS projections. Numbers indicated in brackets are the typical values when applying VGG-VD \cite{simonyan2014very} with the selected pooling method on a 1000-class classification task. The output dimension of our compact bilinear feature is 2 orders of magnitude smaller than the bilinear feature dimension. In practice, the proposed compact representations achieve similar performance to the fully bilinear representation using only $2\%$ of the bilinear feature dimension, suggesting a remarkable $98\%$ redundancy in the bilinear representation.

The RM projection requires moderate amounts of parameter memory (i.e. the random generated but fixed matrix), while TS require almost no parameter memory. If a linear classifier is used after the pooling layer, i.e, a fully connected layer followed by a softmax loss, the number of classifier parameters increases linearly with the pooling output dimension and the number of classes. In the case mentioned above, classification parameters for bilinear pooling would require $1000$MB of storage. Our compact bilinear method, on the other hand, requires far fewer parameters in the classification layer, potentially reducing the risk of over-fitting, and performing better in few shot learning scenarios~\cite{fei2006one},  or domain adaptation~\cite{daume2009frustratingly} scenarios.

Computationally, Tensor Sketch is linear in $d \log d + c$, whereas bilinear is quadratic in $c$, and Random Maclaurin is linear in $cd$ (\tblref{theory_compare}). In practice, the computation time of the pooling layers is dominated by that of the convolution layers. With the Caffe implementation and K40c GPU, the forward backward time of the 16-layer VGG \cite{simonyan2014very} on a $448 \times 448$ image is 312ms. Bilinear pooling requires 0.77ms and TS (with $d = 4096$) requires 5.03ms . TS is slower because FFT has a larger constant factor than matrix multiplication.

\subsection{Alternative dimension reduction methods}
\label{sec:alternative_dim_reduction}
PCA, which is a commonly used dimensionality reduction method, is not a viable alternative in this scenario due to the high dimensionality of the bilinear feature. Solving a PCA usually involves operations on the order of $O(d^3)$, where $d$ is the feature dimension. This is impractical for the high dimensionality, $d=262$K used in bilinear pooling.

Lin et al.~\cite{lin2015bilinear} circumvented these limitations by using PCA \emph{before} forming the bilinear feature, reducing the bilinear feature dimension on CUB200~\cite{welinder2010caltech} from 262,000 to 33,000. While this is a substantial improvement, it still accounts for $12.6\%$ of the original dimensionality. Moreover, the PCA reduction technique requires an expensive initial sweep over the whole dataset to get the principle components. In contrast, our proposed compact bilinear methods do not require any pre-training and can be as small as 4096 dimensions. For completeness, we compare our method to this baseline in Section \ref{sec:pca_bilinear_baseline}.

Another alternative is to use a random projections. However, this requires forming the whole bilinear feature and projecting it to lower dimensional using some random linear operator. Due to the Johnson-Lindenstrauss lemma \cite{dasgupta1999elementary}, the random projection largely preserves pairwise distances between the feature vectors. However, deploying this method requires constructing and storing both the bilinear feature and the fixed random projection matrix. For example, for VGG-VD, the projection matrix will have a shape of $c^2 \times d$, where $c$ and $d$ are the number of channels in the previous layer and the projected dimension, as above. With $d=10,000$ and $c=512$, the projection matrix has $~2.6$ billion entries, making it impractical to store and work with. A classical dense random Gaussian matrix, with entries being i.i.d. $N(0,1)$, would occupy $10.5$GB of memory, which is too much for a high-end GPU such as K40. A sparse random projection matrix would improve the memory consumption to around $40$MB\cite{li2006very}, but would still requires forming bilinear feature first. Furthermore, it requires sparse matrix operations on GPU, which are inevitably slower than dense matrix operations, such as the one used in RM (\algref{RM}).

\squeeze
\section{Experiments}
\label{sec:experiments}
\squeeze
In this section we detail four sets of experiments. First, in \secref{dims_matters}, we investigate some design-choices of the proposed pooling methods: appropriate dimensionality, $d$ and whether to tune the projection parameters, $W$. Second, in \secref{pca_bilinear_baseline}, we conduct a baseline comparison against a PCA based compact pooling method. Third, in \secref{dataset_comparison}, we look at how bilinear pooling in general, and the proposed compact methods in particular, perform in comparison to state-of-the-art on three common computer vision benchmark data-sets. Fourth, in \secref{applications}, we investigate a situation where a low-dimensional representation is particularly useful: few-shot learning. We begin by providing the experimental details.

\subsection{Experimental details}
We evaluate our design on two network structures: the M-net in \cite{chatfield2014return} (VGG-M) and the D-net in \cite{simonyan2014very} (VGG-D). We use the convolution layers of the each network as the local descriptor extractor. More precisely, in the notation of \secref{methods}, $x_s$ is the activation at each spatial location of the convolution layer output. Specifically, we retain the first 14 layers of VGG-M (conv$_5$ + ReLU) and the first 30 layers in VGG-D (conv$_{5\_3}$ + ReLU), as used in \cite{lin2015bilinear}. In addition to bilinear pooling, we also compare to fully connected layer and improved fisher vector encoding \cite{perronnin2010improving}. The latter one is known to outperform other clustering based coding methods\cite{cimpoi2015deep}, such as hard or soft vector quantization \cite{leung2001representing} and VLAD \cite{jegou2010aggregating}. All experiments are performed using MatConvNet~\cite{vedaldi15matconvnet}, and we use $448 \times 448$ input image size, except fully connected pooling as mentioned below. 
\squeeze
\subsubsection{Pooling Methods}
\squeeze
\paragraph{ }
\textbf{Full Bilinear Pooling:} Both VGG-M and VGG-D have $512$ channels in the final convolutional layer, meaning that the bilinear feature dimension is $512\times 512 \approx 250$K. We use a symmetric underlying network structure, corresponding to the B-CNN[M,M] and B-CNN[D,D] configurations in \cite{lin2015bilinear}. We did not experiment with the asymmetric structure such as B-CNN[M, D] because it is shown to have similar performance as the B-CNN[D,D] \cite{lin2015bilinear}. Before the final classification layer, we add an element-wise signed square root layer ($y=\textrm{sign}(x)\sqrt{|x|}$) and an instance-wise $\ell_2$ normalization.

\textbf{Compact Bilinear Pooling:} Our two proposed compact bilinear pooling methods are evaluated in the same exact experimental setup as the bilinear pooling, including the signed square root layer and the $\ell_2$ normalization layer. Both compact methods are parameterized by a used-defined projection dimension $d$ and a set of random generated projection parameters. For notational convenience, we use $W$ to refer to the projection parameters, although they are generated and used differently (Algs. \ref{alg:RM}, \ref{alg:TS}). When integer constraints are relaxed, $W$ can be learned as part of the end-to-end back-propagation. The appropriate setting of $d$, and of whether or not to tune $W$, depends on the amount of training data, memory budget, and the difficulty of the classification task. We discuss these design choices in \secref{dims_matters}; in practice we found that $d = 8000$ is sufficient for reaching close-to maximum accuracy, and that tuning the projection parameters has a positive, but small, boost.

\textbf{Fully Connected Pooling:} The fully connected baseline refer to a classical fine tuning scenario, where one starts from a network trained on a large amount of images, such as VGG-M, and replace the last classification layer with a random initialized $k$-way classification layer before fine-tuning. We refer to this as the "fully connected" because this method has two fully connected layers between the last convolution layer and the classification layer.  This method requires a fixed input image sizes, dictated by the network structure. For the VGG nets used in this work, the input size is $224 \times 224$, and we thus re-size all images to this size for this method.

\textbf{Improved Fisher Encoding:} Similarly to bilinear pooling, fisher encoding~\cite{perronnin2010improving} has recently been used as an encoding \& pooling alternative to the fully connected layers~\cite{cimpoi2015deep}. Following~\cite{cimpoi2015deep}, the activations of last convolutional layer (excluding ReLU) are used as input the encoding step, and the encoding uses 64 GMM components. 

\squeeze
\subsubsection{Learning Configuration}
\squeeze
During fine-tuning, we initialized the last layer using the weights of the trained logistic regression and attach a corresponding logistic loss. We then fine tune the whole network until convergence using a constant small learning rate of $10^{-3}$, a weight decay of $5\times 10^{-4}$, a batch size of 32 for VGG-M and 8 for VGG-D. In practice, convergence occur in $<100$ epochs. Note that for RM and TS, back-propagation can be used simply as a way to tune the deeper layers of the network (as it is used in full bilinear pooling), or to also tune the projection parameters, $W$. We investigate both options in \secref{dims_matters}. Fisher vector has an unsupervised dictionary learning phase, and it is unclear how to perform fine-tuning~\cite{cimpoi2015deep}. We therefore do not evaluate Fisher Vector under fine-tuning.

In \secref{dims_matters} we also evaluate each method as a feature extractor. Using the forward-pass through the network, we train a linear classifier on the activations. We use $\ell_2$ regularized logistic regression: $\lambda ||w||_2^2 + \sum_i l(\langle x_i, w \rangle, y_i)$ with $\lambda=0.001$ as we found that it slightly outperforms SVM.


\subsection{Configurations of compact pooling}
\label{sec:dims_matters}
Both RM and TS pooling have a user defined projection dimension $d$, and a set of projection parameters, $W$. To investigate the parameters of the proposed compact bilinear methods, we conduced extensive experiments on the CUB-200 \cite{wah2011caltech} dataset which contains 11,788 images of 200 bird species, with a fixed training and testing set split. We evaluate in the mode where part annotations are not provided at neither training nor testing time, and use VGG-M for all experiments in this section. 

\begin{figure}[t]
\begin{center}
   \includegraphics[width=\linewidth]{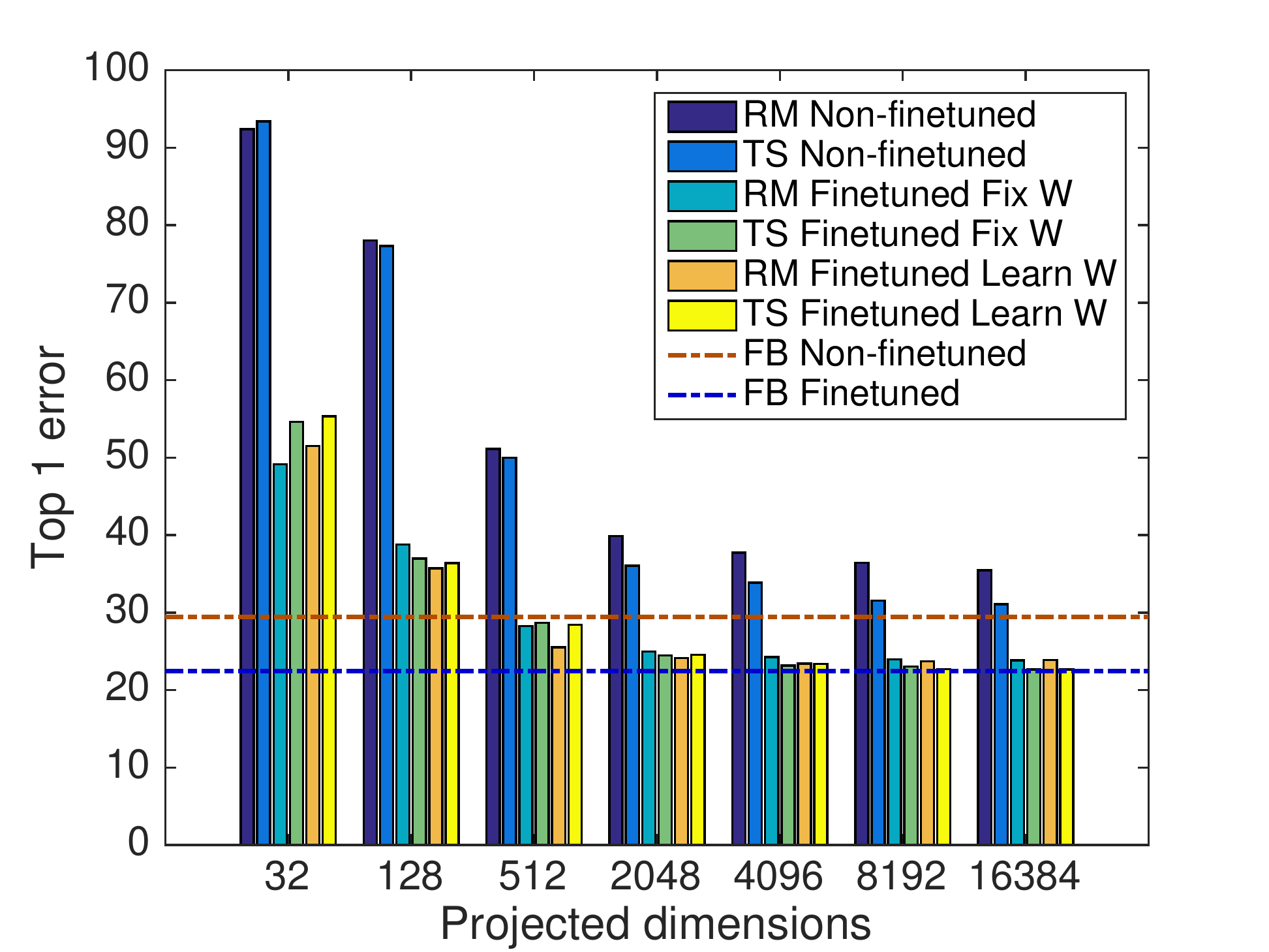}
\end{center}
   \caption{Classification error on the CUB dataset. Comparison of Random Maclaurin (RM) and Tensor Sketch (TS) for various combinations of projection dimensions and fine-tuning options. The two horizontal lines shows the performance of fine-tuned and non fine-tuned Fully Bilinear (FB).}
\label{fig:dim} \squeeze
\vskip -0.5cm
\end{figure}

\figref{dim} summarizes our results. As the projection dimension $d$ increases, the two compact bilinear methods reach the performance of the full bilinear pooling. When not fine-tuned, the error of TS with $d=16$K is $1.7\%$ less than that of bilinear feature, while only using $6.1\%$ of the original number of dimensions. When fine tuned, the performance  gap disappears: TS with $d=16$K has an error rate of $22.66\%$, compared to $22.44\%$ of bilinear pooling. 

In lower dimension, RM outperforms TS, especially when tuning $W$. This may be because RM pooling has more parameters, which provides additional learning capacity despite the low-dimensional output (\tblref{theory_compare}). Conversely, TS outperforms RM when $d>2000$. This is consistent with the results of Pham \& Pagm, who evaluated these projections methods on several smaller data-sets~\cite{pham2013fast}. Note that these previous studies did not use pooling nor fine-tuning as part of their experimentation.

\figref{dim} also shows performances using extremely low dimensional representation, $d=32, 128$ and $512$. While the performance decreased significantly for the fixed representation, fine-tuning brought back much of the discriminative capability. For example, $d=32$ achieved less than $50\%$ error on the challenging 200-class fine grained classification task. Going up slightly, to $512$ dimensions, it yields $25.54\%$ error rate. This is only $3.1\%$ drop in performance compared to the 250,000 dimensional bilinear feature. Such extremely compact but highly discriminative image feature representations are useful, for example, in image retrieval systems. For comparison, Wang et al. used a 4096 dimensional feature embedding in their recent retrieval system~\cite{wang2014learning}. 

In conclusion, our experiments suggest that between $2000$ and $8000$ features dimension is appropriate. They also suggest that the projection parameters, $W$ should only be tuned when one using extremely low dimensional representations (the 32 dimensional results is an exception). Our experiments also confirmed the importance of fine-tuning, emphasizing the critical importance of using projection methods which allow fine-tuning.

\subsection{Comparison to the PCA-Bilinear baseline}
\label{sec:pca_bilinear_baseline}
As mentioned in Section \ref{sec:alternative_dim_reduction}, a simple alternative dimensionality reduction method would be to use PCA before bilinear pooling~\cite{lin2015bilinear}. We compare this approach with our compact Tensor Sketch method on the CUB\cite{wah2011caltech} dataset with VGG-M \cite{chatfield2014return} network. The PCA-Bilinear baseline is implemented by inserting an $1 \times 1$ convolution before the bilinear layer with weights initialized by PCA. The number of outputs, $k$ of this convolutional layer will determine the feature dimension ($k^2$). 

Results with various $k^2$ are shown in \tblref{pca}. The gap between the PCA-reduced bilinear feature and TS feature is large especially when the feature dimension is small and network not fine tuned. When fine tuned, the gap shrinks but the PCA-Bilinear approach is not good at utilizing larger dimensions. For example, the PCA approach reaches a 23.8\% error rate at 16K dimensions, which is larger than the 23.2\% error rate of TS at 4K dimensions. 
\begin{table}[htb]
\small
\centering
\begin{tabular}{l|llll}
\hline
 dim.    & 256 & 1024 & 4096 & 16384 \\
\hline \hline
PCA & 72.5/42.9    &  49.7/28.9  &  41.3/25.3  &   36.2/23.8    \\
TS   & 62.6/32.2  &  41.6/25.5 & 33.9/23.2 & 31.1/22.5      \\
\hline    
\end{tabular}
\caption{Comparison between PCA reduced feature and TS. Numbers refer to Top 1 error rates without and with fine tuning respectively. } 
\label{tbl:pca}
\vskip -0.3cm
\end{table}

\subsection{Evaluation across multiple data-sets}
\label{sec:dataset_comparison}
Bilinear pooling has been studied extensively. Carreira et al. used second order pooling to facilitate semantic segmentation~\cite{carreira2012semantic}. Lin et al. used bilinear pooling for fine-grained visual classification~\cite{lin2015bilinear}, and Rowchowdhury used bilinear pooling for face verification~\cite{roychowdhury2015face}. These methods all achieved state-of-art on the respective tasks indicating the wide utility of bilinear pooling. In this section we show that the compact representations perform on par with bilinear pooling on three very different image classification tasks. Since the compact representation requires orders of magnitude less memory, this suggests that it is the preferable method for a wide array of visual recognition tasks.

\begin{figure}[htb]
  \centering
\begin{subfigure}{0.3\linewidth}
  \centering
  \includegraphics[width=.95\linewidth, height=2cm]{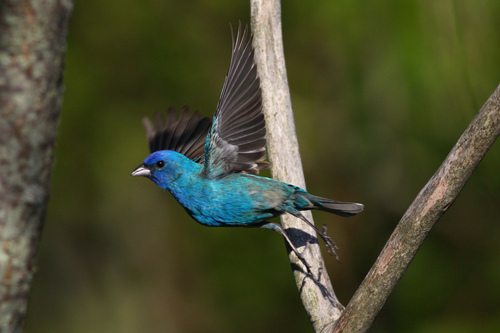}
  \label{fig:sfig1}
\end{subfigure}%
\begin{subfigure}{0.3\linewidth}
  \centering
  \includegraphics[width=.95\linewidth, height=2cm]{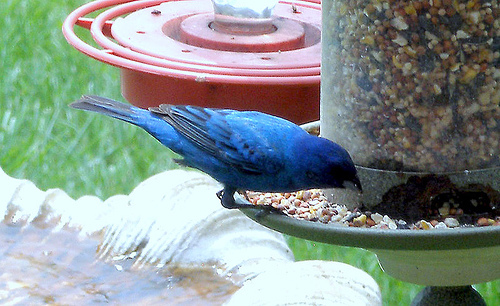}
  \label{fig:sfig2}
\end{subfigure}%
\begin{subfigure}{0.3\linewidth}
  \centering
  \includegraphics[width=.95\linewidth, height=2cm]{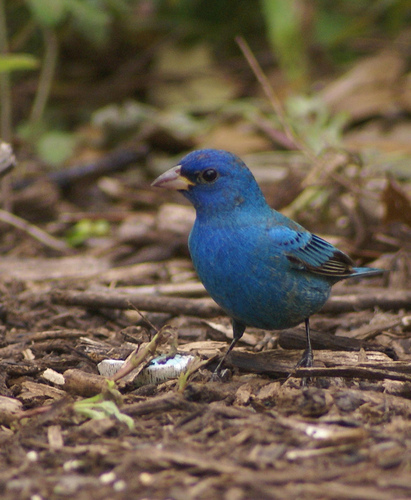}
  \label{fig:sfig2}
\end{subfigure}
\vskip -0.3cm
\begin{subfigure}{0.3\linewidth}
  \centering
  \includegraphics[width=.95\linewidth, height=2cm]{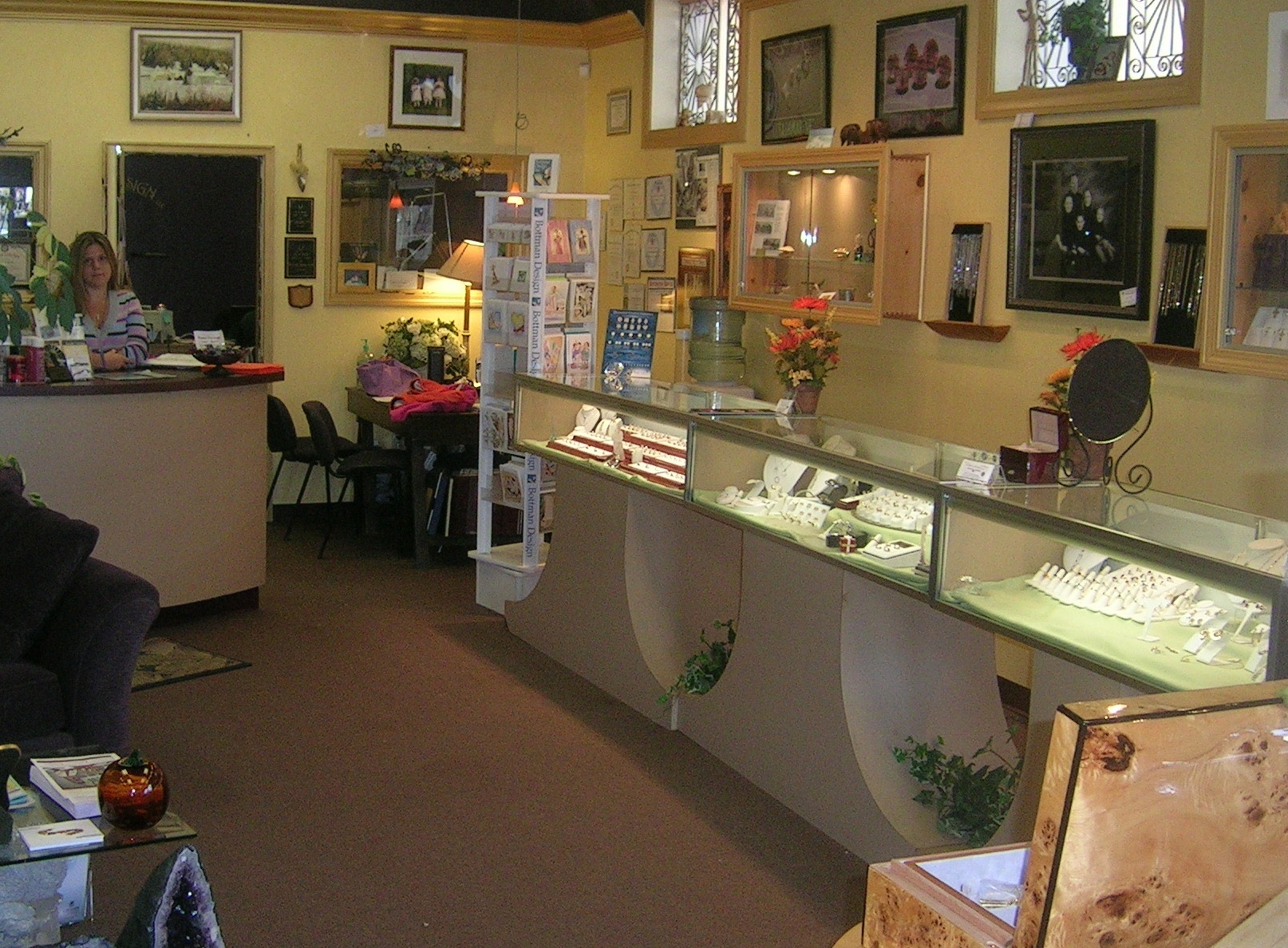}
  \label{fig:sfig1}
\end{subfigure}%
\begin{subfigure}{0.3\linewidth}
  \centering
  \includegraphics[width=.95\linewidth, height=2cm]{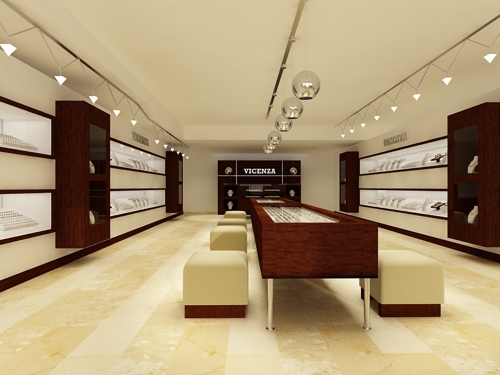}
  \label{fig:sfig2}
\end{subfigure}%
\begin{subfigure}{0.3\linewidth}
  \centering
  \includegraphics[width=.95\linewidth, height=2cm]{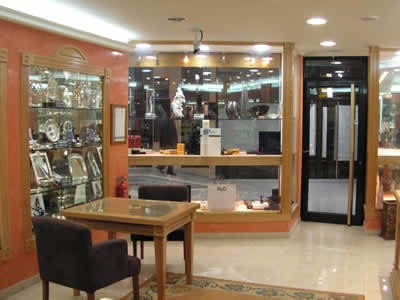}
  \label{fig:sfig2}
\end{subfigure}
\vskip -0.3cm
\begin{subfigure}{0.3\linewidth}
  \centering
  \includegraphics[width=.95\linewidth, height=2cm]{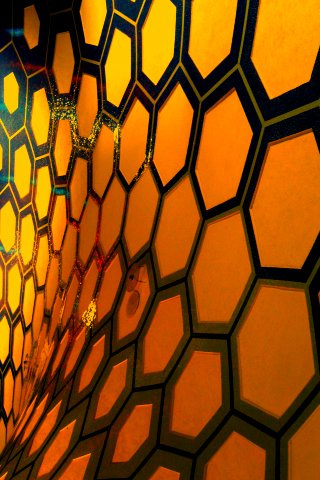}
  \label{fig:sfig1}
\end{subfigure}%
\begin{subfigure}{0.3\linewidth}
  \centering
  \includegraphics[width=.95\linewidth, height=2cm]{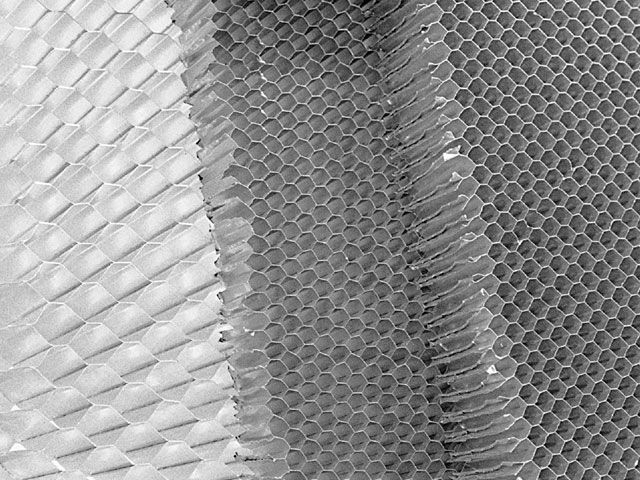}
  \label{fig:sfig2}
\end{subfigure}%
\begin{subfigure}{0.3\linewidth}
  \centering
  \includegraphics[width=.95\linewidth, height=2cm]{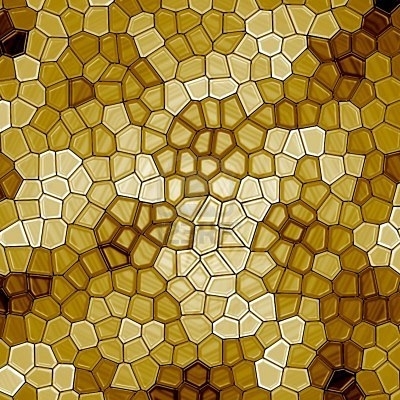}
  \label{fig:sfig2}
\end{subfigure}
\vskip -0.2cm
   \caption{Samples images from the three datasets examined in \secref{dataset_comparison}. Each row contains samples from indigo bunting in CUB, jewelery shop in MIT, and honey comb in DTD.}
\vskip -0.5cm
\label{fig:sample_images}
\end{figure}

Fully connected pooling, fisher vector encoding, bilinear pooling and the two compact bilinear pooling methods are compared on three visual recognition tasks: fine-grained visual categorization represented by CUB-200-2011~\cite{wah2011caltech}, scene recognition represented by the MIT indoor scene recognition dataset ~\cite{quattoni2009recognizing}, and texture classification represented by the Describable Texture Dataset ~\cite{cimpoi14describing}. Sample figures are provided in \figref{sample_images}, and dataset details in \tblref{dataset_details}. Guided by our results in \secref{dims_matters} we use $d=8192$ dimensions and fix the projection parameters $W$.

\begin{table*}[htb]
\small
\begin{center}
\begin{tabular}{l|l|l|l|l|l|l}
\hline
\textbf{Data-set} & \textbf{Net}  & \textbf{FC}~\cite{chatfield2014return,simonyan2014very} & \textbf{Fisher}~\cite{cimpoi2015deep} & \textbf{FB}~\cite{lin2015bilinear} & \textbf{RM} (\algref{RM}) & \textbf{TS} (\algref{TS})\\
\hline       
\hline       
CUB~\cite{wah2011caltech} & VGG-M~\cite{chatfield2014return} & 49.90/42.03   & 52.73/NA    & 29.41/\textbf{22.44} & 36.42/23.96 & 31.53/23.06   \\

CUB~\cite{wah2011caltech} & VGG-D~\cite{simonyan2014very} & 42.56/33.88   & 35.80/NA    & 19.90/\textbf{16.00} & 21.83/16.14 & 20.50/\textbf{16.00}   \\
MIT~\cite{quattoni2009recognizing} & VGG-M~\cite{chatfield2014return} & 39.67/35.64   & 32.80/NA    & \textbf{29.77}/32.95 & 31.83/32.03 & 30.71/31.30 \\
MIT~\cite{quattoni2009recognizing} & VGG-D~\cite{simonyan2014very} & 35.49/32.24	& 24.43/NA &	\textbf{22.45}/28.98$^*$ & 26.11/26.57 &	23.83/27.27$^*$ \\
DTD~\cite{cimpoi14describing} & VGG-M~\cite{chatfield2014return}  & 46.81/43.22	& 42.58/NA	& \textbf{39.57}/40.50 &	43.03/41.36	& \textbf{39.60}/40.71   \\
DTD~\cite{cimpoi14describing} & VGG-D~\cite{simonyan2014very} & 39.89/40.11 &	34.47/NA &	32.50/35.04 &	36.76/34.43 &	\textbf{32.29}/35.49 \\
\hline
\end{tabular}
\end{center}
\caption{Classification error of fully connected (FC), fisher vector, full bilinear (FB) and compact bilinear pooling methods, Random Maclaurin (RM) and Tensor Sketch (TS). For RM and TS we set the projection dimension, $d=8192$ and we fix the projection parameters, $W$. The number before and after the slash represents the error without and with fine tuning respectively. Some fine tuning experiments diverged, when VGG-D is fine-tuned on MIT dataset. These are marked with an asterisk and we report the error rate at the 20th epoch.}
\label{tbl:results_datasets} \squeeze
\end{table*}

\begin{table}[htb]
\small
\centering \begin{tabular}{l|ccc}
\hline
Data-set            & \# train img & \# test img & \# classes \\
\hline\hline
CUB~\cite{wah2011caltech} & 5994 & 5794 & 200 \\
MIT~\cite{quattoni2009recognizing} & 4017 & 1339 & 67 \\
DTD~\cite{cimpoi14describing} & 1880 & 3760 & 47 \\
\hline    
\end{tabular}
\caption{Summary statistics of data-sets in ~\secref{dataset_comparison}} \label{tbl:dataset_details} \squeeze
\vskip -0.3cm
\end{table}

\squeeze
\subsubsection{Bird species recognition}
\squeeze
CUB is a fine-grained visual categorization dataset. Good performance on this dataset requires identification of overall bird shape, texture and colors, but also capacity to focus on subtle differences, such as the beak-shapes. The only supervision we use is the image level class labels, without referring to either part or bounding box annotations.

Our results indicate that bilinear and compact bilinear pooling outperforms fully connected and fisher vector by a large margin, both with and without fine-tuning (\tblref{results_datasets}). Among the compact bilinear methods, TS consistently outperformed RS. For the larger VGG-D network, bilinear pooling achieved 19.90\% error rate before fine tuning, while RM and TS achieved 21.83\% and 20.50\% respectively. This is a modest 1.93\% and 0.6\% performance loss considering the huge reduction in feature dimension (from $250k$ to $8192$). Notably, this difference disappeared after fine-tuning when the bilinear pooling methods all reached an error rate of $~16.0\%$. This is, to the best of our knowledge, the state of the art performance on this dataset without part annotation~\cite{krause2015fine,lin2015bilinear}. The story is similar for the smaller VGG-M network: TS is more favorable than RM and the performance gap between compact full bilinear shrinks to $0.5\%$ after fine tuning. 

\subsubsection{Indoor scene recognition}
Scene recognition is quite different from fine-grained visual categorization, requiring localization and classification of discriminative and non-salient objects. As shown in \figref{sample_images}, the intra-class variation can be quite large.

As expected, and previously observed~\cite{cimpoi2015deep}, improved Fisher vector encoding outperformed fully connected pooling by 6.87\% on the MIT scene data-set (\tblref{results_datasets}). More surprising, bilinear pooling outperformed Fisher vector by 3.03\%. Even though bilinear pooling was proposed for object-centric tasks, such as fine grained visual recognition, this experiment thus suggests that is it appropriate also for scene recognition. Compact TS performs slightly worse (0.94\%) than full bilinear pooling, but 2.09\% better than Fisher vector. This is notable, since fisher vector is used in the current state-of-art method for this dataset~\cite{cimpoi2015deep}. 

Surprisingly, fine-tuning negatively impacts the error-rates of the full and compact bilinear methods, about 2\%. We believe this is due to the small training-set size and large number of convolutional weights in VGG-D, but it deserves further attention. 

\subsubsection{Texture classification}
Texture classification is similar to scene recognition in that it requires attention to small features which can occur anywhere in the image plane. Our results confirm this, and we see similar trends as on the MIT data-set (Table \ref{tbl:results_datasets}). Again, Fisher encoding outperformed fully connected pooling by a large margin and RM pooling performed on par with Fisher encoding, achieving $\sim 34.5\%$ error-rate using VGG-D. Both are out-performed by $\sim 2\%$ using full bilinear pooling which achieves $32.50\%$. The compact TS pooling method achieves the strongest results at $32.29\%$ error-rate using the VGG-D network. This is $2.18\%$ better than the fisher vector and the lowest reported single-scale error rate on this data-set\footnote{Cimpoi et al. extract descriptors at several scales to achieve their state-of-the-art results~\cite{cimpoi2015deep}}. Again, fine-tuning did not improve the results for full bilinear pooling or TS, but it did for RM.

\subsection{An application to few-shot learning}
\label{sec:applications}

Few-shot learning is the task of generalizing from a very small number of labeled training samples~\cite{fei2006one}. It is important in many deployment scenarios where labels are expensive or time-consuming to acquire~\cite{beijbom2012automated}.

Fundamental results in learning-theory show a relationship between the number of required training samples and the size of the hypothesis space (VC-dimension) of the classifier that is being trained~\cite{vapnik2013nature}. For linear classifiers, the hypothesis space grows with the feature dimensions, and we therefore expect a lower-dimensional representation to be better suited for few-shot learning scenarios. We investigate this by comparing the full bilinear pooling method ($d=250,000$) to TS pooling ($d=8192$). For these experiments we do not use fine-tuning and use VGG-M as the local feature extractor.

When only one example is provided for each class, TS achieves a score of 15.5\%, which is a 22.8\% relative improvement over full bilinear pooling, or 2.9\% in absolute value, confirming the utility of a low-dimensional descriptor for few-shot learning. The gap remains at 2.5\% with 3 samples per class or 600 training images. As the number of shots increases, the scores of TS and the bilinear pooling increase rapidly, converging around 15 images per class, which is roughly half the dataset.  (\tblref{dataset_details}). 
 
\begin{table}[htb]
\small
\begin{center}
\begin{tabular}{l|cccccc} 
\hline
\# images & 1 & 2 & 3 & 7 & 14 \\
\hline
\hline
Bilinear & 12.7 & 21.5 & 27.4 & 41.7 & 53.9   \\
TS 	& 15.6 & 24.1 & 29.9 & 43.1 & 54.3 &  \\
\hline
\end{tabular}
\end{center}
\vspace{-3mm}
\caption{Few shot learning comparison on the CUB data-set. Results given as mean average precision for $k$ training images from \emph{each} of the $200$ categories.}
\squeeze
\label{tbl_few_shots}
\squeeze
\vskip -0.3cm
\end{table}

\squeeze
\section{Conclusion}
We have modeled bilinear pooling in a kernelized framework and suggested two compact representations, both of which allow back-propagation of gradients for end-to-end optimization of the classification pipeline. Our key experimental results is that an 8K dimensional TS feature has the same performance as a 262K bilinear feature, enabling a remarkable $96.5\%$ compression. TS is also more compact than fisher encoding, and achieves stronger results. We believe TS could be useful for image retrieval, where storage and indexing are central issues or in situations which require further processing: e.g. part-based models~\cite{branson2013efficient,felzenszwalb2010object}, conditional random fields, multi-scale analysis, spatial pyramid pooling or hidden Markov models; however these studies are left to future work.  Further, TS reduces network and classification parameters memory significantly which can be critical e.g. for deployment on embedded systems. Finally, after having shown how bilinear pooling uses a pairwise polynomial kernel to compare local descriptors, it would be interesting to explore how alternative kernels can be incorporated in deep visual recognition systems.

{\small
\bibliographystyle{ieee}
\bibliography{compact_bilinear_pooling}
}

\end{document}